%% file: acl2024.tex
\pdfoutput=1

\documentclass[11pt]{article}

\usepackage[]{ACL2024}

\usepackage{times}
\usepackage{latexsym}

\usepackage{algorithm}
\usepackage{multirow}
\usepackage{graphicx}
\usepackage{subfigure}
\usepackage{amsmath}
\usepackage{enumitem}
\usepackage{booktabs}
\usepackage{xspace}
\usepackage{microtype}
\usepackage{multicol}
\usepackage{caption}
\usepackage{amssymb}
\usepackage{array}
\usepackage{fp}
\usepackage{makecell}
\usepackage{flushend}
\usepackage{cases}
\usepackage{color}
\usepackage{algpseudocode}
\usepackage{listings}
\usepackage{xcolor}
\usepackage{mathrsfs}
\usepackage{pifont}

\newcommand{\paratitle}[1]{\vspace{1.5ex}\noindent\textbf{#1}}
\newcommand{\ie}{\emph{i.e.,}\xspace}

\newcommand{\eg}{\emph{e.g.,}\xspace}

\newcommand{\ignore}[1]{}

\usepackage[T1]{fontenc}

\usepackage[utf8]{inputenc}

\usepackage{microtype}

\usepackage{inconsolata}

\title{RAG-Star: Enhancing Deliberative Reasoning with Retrieval Augmented Verification and Refinement}

\author{
    \textbf{Jinhao Jiang\textsuperscript{{1}}\thanks{\llap{}\:\:\:Equal contributions.},
            Jiayi Chen\textsuperscript{{2}}\footnotemark[1],
            Junyi Li\textsuperscript{{4}}\footnotemark[1],
            Ruiyang Ren\textsuperscript{{1}},
            Shijie Wang\textsuperscript{{3}}}\\
    \textbf{Wayne Xin Zhao\textsuperscript{{1}}\thanks{\llap{}\:\:\:Corresponding author.},
            Yang Song\textsuperscript{{5}}\footnotemark[2],
            Tao Zhang\textsuperscript{{5}}}\\
	\textsuperscript{1}Gaoling School of Artificial Intelligence, Renmin University of China.\\
	\textsuperscript{2}Wuhan University of Science and Technology.
    \textsuperscript{3}Northeastern University at Qinhuangdao.\\
	\textsuperscript{4}Department of Computer Science, National University of Singapore. \textsuperscript{5}BOSS Zhipin, Beijing, China.\\
    \texttt{jiangjinhao@ruc.edu.cn, batmanfly@gmail.com}\\
}

\begin{document}
\maketitle

\begin{abstract}


Existing large language models (LLMs) show exceptional problem-solving capabilities but might struggle with complex reasoning tasks. Despite the successes of chain-of-thought and tree-based search methods, they mainly depend on the internal knowledge of LLMs to search over intermediate reasoning steps, limited to dealing with simple tasks involving fewer reasoning steps. In this paper, we propose \textbf{RAG-Star}, a novel RAG approach that integrates the retrieved information to guide the tree-based deliberative reasoning process that relies on the inherent knowledge of LLMs. By leveraging Monte Carlo Tree Search, RAG-Star iteratively plans intermediate sub-queries and answers for reasoning based on the LLM itself. To consolidate internal and external knowledge, we propose an retrieval-augmented verification that utilizes query- and answer-aware reward modeling to provide feedback for the inherent reasoning of LLMs. Our experiments involving Llama-3.1-8B-Instruct and GPT-4o demonstrate that RAG-Star significantly  outperforms previous RAG and reasoning methods.

\end{abstract}

\input{sec-intro}

\input{sec-related_work}

\input{sec-preliminary}

\input{sec-approach}

\input{sec-experiment}

\input{sec-conclusion}

\input{sec-limitation}

\bibliography{custom}
\bibliographystyle{acl_natbib}


\clearpage

\end{document}

%% file: sec-intro.tex
\section{Introduction}
\label{introduction}
Despite the excellent capabilities of large language models (LLMs)~\citep{llm_survey}, they still 
face significant challenges in complex reasoning tasks (\eg multi-hop question answering),  which often go beyond simple, single-step problem-solving, demanding a deeper level of cognitive reasoning across multiple facts, sources, or contexts~\citep{abs-2406-12753,bbh}. Great efforts have been made to improve the reasoning effectiveness of LLMs by conducing step-by-step reasoning, exemplified by chain-of-thought (CoT)~\citep{cot}. 
However, as the number of reasoning steps grows, LLMs are often prone to introduce logical errors, factual hallucinations, or  inconsistent statements~\citep{cot,LyuHSZRWAC23}. 

\ignore{
\textcolor{blue}{Recent research has attributed the above issue to the limitation of auto-regressive generation and the absence of inherent knowledge of LLMs~\citep{liu2024much,abs-2406-14283,li2024can}.  Although extensive research has sought to 
augment LLMs with external information sources (\emph{a.k.a.} retrieval-augmented generation, RAG)~\citep{LewisPPPKGKLYR020,Yao-arxiv-2022-ReAct}, external knowledge is treated as direct input to LLMs via prompts, which might lead to conflicts between internal and external knowledge and still remains constrained by the auto-regressive nature}.
}
In fact,  step-by-step reasoning in the auto-regressive generation paradigm can be described as akin to ``{System 1}'', a mode of thinking which is fast, instinctive but less accurate~\citep{kahneman2011thinking}. 
Conversely, solving complex reasoning problems requires more in-depth, deliberative, and logical thinking, known as the ``{System 2}'' mode, which requires conscious effort to conduct massive strategic decision-making~\citep{kahneman2011thinking}. To enhance the ``System 2'' reasoning capabilities of LLMs, prior studies have proposed to conduct deliberative generation by leveraging basic tree search algorithms (\eg 
Monte Carlo Tree Search~\citep{alphazero}).
However, LLMs in these studies mainly depend on their \emph{internal knowledge} to search over intermediate reasoning steps, limited to handling problems with relatively simple reasoning process.
To leverage external knowledge in model reasoning, extensive research has sought to 
augment LLMs with external information sources (\emph{a.k.a.} retrieval-augmented generation, RAG)~\citep{LewisPPPKGKLYR020,Yao-arxiv-2022-ReAct}, while existing efforts mainly consider sequential reasoning structure, which cannot naturally support more complex reasoning structure like MCTS. 
Thus, we raise the following research question: \emph{Can RAG enhance the deliberative reasoning capabilities of LLMs?}   


In light of this, in this paper, we propose \textbf{RAG-Star}, a novel RAG-enhanced framework designed to improve multi-step reasoning capabilities of LLMs with deliberative planning. 
As the major technical contribution, RAG-Star can fully exploit the internal knowledge of LLMs to plan the multi-step reasoning, and meanwhile integrating the external retrieval to guide the internal reasoning process. 
To achieve this goal, 
we first introduce a tree-based search algorithm (\ie Monte Carlo Tree Search, MCTS) with LLMs to search over possible plans for solving the problem at hand where a complete plan is composed of a sequence of sub-queries and corresponding answers. Starting from the input question (root node), RAG-Star iteratively generates and selects an appropriate sub-query and its answer (intermediate node), which aims to maximally explore the optimal sub-query path towards the final answer solely based on the inherent knowledge of LLMs.  
Second, different from existing deliberation methods~\citep{abs-2406-14283,Yao-arxiv-2023-Tree}, RAG-Star proposes retrieval-augmented verification that involves both query- and answer-aware reward modeling, fully exploiting external sources to guide the internal deliberative reasoning. 
In our approach, instead of directly interfering in the reasoning process of LLMs, we consider employing RAG to refine the derived reasoning steps in MCTS, which can effectively reduce the conflicts between inherent and external knowledge, which has been a common issue when using RAG methods~\citep{wang2024astute,gao2023retrieval}.



We conduct extensive experiments to verify the effectiveness of RAG-Star based on Llama-3.1-8B-Instruct and GPT-4o.
Our method outperforms the baselines by up to 18.98\% and 16.19\% on average for Llama-3.1-8B and GPT-4o, respectively. 


Our main contributions can be summarized as:

$\bullet$ We propose RAG-Star that leverages external retrieval to enahnce the deliberative reasoning of LLMs based on their internal knowledge.

$\bullet$ We design an effective retrieval-augmented verification and refinement to evaluate and correct the inherent reasoning process.

$\bullet$ We conduct extensive experiments on several datasets, where RAG-Star significantly outperforms existing RAG and reasoning methods.

%% file: sec-related_work.tex
\section{Related Work}
\paratitle{Retrieval-Augmented LLMs.} 
Augmenting large language models (LLMs) with retrieval has been extensively studied in existing literature~\cite{rag,borgeaud2022improving,realm}, which incorporates a differentiable retriever to provide external sources for LLMs. Furthermore, LLMs have made significant advancements in many reasoning tasks, such as code generation~\citep{OpenAI-OpenAI-2023-GPT-4}, math word problems~\citep{ZhuWZZ0GZY23} and question answering~\citep{gpt3}. Chain-of-thought (CoT) has been reported as an emergent ability of LLMs when they are large enough~\citep{cot}, which encourages LLMs to generate explicit intermediate reasoning steps in reasoning rather than simply providing answers directly. To elicit or improve the multi-step reasoning capability of LLMs, several approaches seek to harness the strengths of both CoT and retrieval on knowledge-intensive complex reasoning tasks, such as multi-hop question answering~\citep{Yao-arxiv-2022-ReAct,ZhaoLJQB23}. 
The rationales gained from reasoning enhance the retrieval of more relevant information, while the retrieved knowledge improves the factuality of intermediate reasoning steps. However, these approaches primarily take retrieved documents as direct input to the model, easily suffering from knowledge conflicts between the parametric knowledge of LLMs and the external sources. 
In contrast, our RAG-Star framework integrates tree-based search to fully explore the solution space and repurpose the retrieval information as external guidance to the reasoning process.


\paratitle{Enhancing LLMs with Search.} Applying search on top of LLMs has been a topic of much interest. Several recent works have explored search algorithms to improve the performance of LLMs during the inference stage~\citep{abs-2406-14283,zhang2024llama}. The bitter lesson~\citep{sutton2019bitter} famously suggests that two forms of scaling, \ie learning and search, supersede all other approaches. Many studies have proven that scaling the inference-time computation can lead to substantial improvements in the performance of LLMs without training~\citep{abs-2407-21787,abs-2408-03314}. 
These search algorithms, where multiple branches of outcomes are explored during search, have been widely applied in reinforcement learning algorithms~\citep{HartNR68,alphazero} and many real-world applications such as AlphaGo~\citep{alphago} for their good exploration-exploitation trade-off. However, these approaches mainly rely on the internal knowledge of LLMs to search potential solutions, which might not be optimal and leads to a amount of rollouts, significantly slowing down the decoding process. 
In this paper, we leverage the external retrieval sources to enhance the deliberative search process with LLMs, effectively differentiate the internal reasoning and external retrieval.

%% file: sec-preliminary.tex
\section{Preliminary}
\label{preliminary}

In this section, we will first formally define our task and then introduce Monte Carlo Tree Search which is used in our proposed RAG-Star approach.

\paratitle{Task Formulation.} 
In this work, we mainly focus on open-domain multi-hop question answering~\citep{multi-hop,hotpotqa}, which requires multiple steps of reasoning across different documents to answer questions. Previous work typically adopts an iterative \emph{reason-then-generate} pipeline~\citep{cot,reasoning_survey}. At each step, the LLM first infers an intermediate sub-query based on the current situation and then generates possible answers to the query.
Formally, given a natural language input question, at the $t$-th step, the LLM $\mathcal{M}_\theta$ (parameterized by $\theta$) first deliberately reasons about a sub-query $q_t$, followed by generating an answer $a_t$ based on its inherent knowledge. In some literature~\citep{Yao-arxiv-2022-ReAct,self-rag}, retrieval-augmented generation (RAG) has been employed to improve the factuality of intermediate reasoning steps. For each sub-query $q_t$, the retriever retrieves top-$K$ documents $\mathcal{D}_t=\{d_{t,k}\}_{k=1}^{K}$ from an external large-scale corpus, \eg Wikipedia, supplying them to the LLM to generate more accurate answers.


\paratitle{Monte Carlo Tree Search~(MCTS).} 
In existing literature~\citep{quiet-star,zhang2024llama}, MCTS builds a search tree $\mathcal{T}$ based on a policy model $\pi_{\theta}$, which is usually the target LLM $\mathcal{M}_\theta$. 
Each node $s_t=[q_t,a_t, N(s_t), V(s_t)]$ represents a state comprising the sub-query $q_t$, its answer $a_t$, the number of visits $N(s_t)$, and the value function (expected reward) $V(s_t)$ for accurately answering questions, except that the root node $s_0 = [q_0]$ only contains the original input question $q_0$, and each edge is an action aiming to generate the next sub-query. 
During the search process, MCTS runs for multiple simulations. For the $t$-th simulation, it conducts four operations to expand the tree: 

$\bullet$ \textit{Selection}  aims to select a node with the highest UCT (Upper Confidence bounds applied to Trees) score~\citep{uct} starting from the root node $s_0$. The UCT score of a child node with state $s_t$ is calculated as follows:
\begin{align}
    \label{eq:uct}
    {UCT}(s_t) = V(s_t) + w\sqrt{\frac{\ln{N(p)}}{N(s_t)}},
\end{align}
where $w$  controls the exploration and exploitation, and $p$ is the parent node of the current  node $s_t$. 

$\bullet$ \textit{Expansion}   explores multiple child nodes $\{s_{t+1}\}$ from the selected node $s_t$ through repeated sampling based on the policy model $\pi_{\theta}$.

$\bullet$ \textit{Simulation}  aims to perform rollout for each expanded child node $s_{t+1}$ until the task is solved and obtain a reward $r$ based on the rollout results. 

$\bullet$ \textit{Backpropagation} operation leverages the reward~$r$ of the child node to update the expected reward $V(s_t)$ of nodes along the path from the root node to the current node:
\begin{align}
    \label{eq:back}
    N_{new}(s_t) &= N_{old}(s_t)+1, \\
    V_{new}(s_t) &= \frac{V_{old}(s_t)N_{old}(s_t)+r}{N_{new}(s_t)},
\end{align}
where $N_{old}(s_t)$ and $V_{old}(s_t)$ are the number of visits and value function at last iteration, respectively.

%% file: sec-approach.tex
\section{Approach}
\label{approach}

\subsection{Overview}
RAG has been an indispensable technique to address the inherent knowledge limitations of LLMs, effectively integrating requisite information and grounding to reliable sources~\citep{rag,realm}.
However, existing work mainly utilizes RAG to provide supplementary knowledge, while overlooking a thorough investigation of RAG on enhancing the inherent reasoning capabilities of LLMs. To address this, we propose {RAG-Star}, a framework to fully harness the potential of internal knowledge in LLMs for multi-step reasoning guided by the external retrieval.

Our RAG-Star framework contains two major technical steps. 
First, we propose \textit{tree-based subquery generation} to perform deliberative reasoning with MCTS, totally relying on the inherent knowledge of LLMs. Second, we design \textit{retrieval-augmented verification} capitalizing on RAG to assist in guiding the reasoning based on the external knowledge. Under this framework, RAG-Star first selects a node from the tree to explore (Section~\ref{sec:method-node-select}), then generates the next sub-query and answers for obtaining new child nodes (Section~\ref{sec:method-plan-expand}), and computes a reward to the expanded nodes (Section~\ref{sec:method-reward-modeling}). Finally, it backpropagates the reward to update their parent nodes on the tree (Section~\ref{sec:method-reward-modeling}). This process will iterate until the task is solved. Next, we will describe each step in detail.


\begin{figure*}[t]
    \centering
    \includegraphics[width=\textwidth]{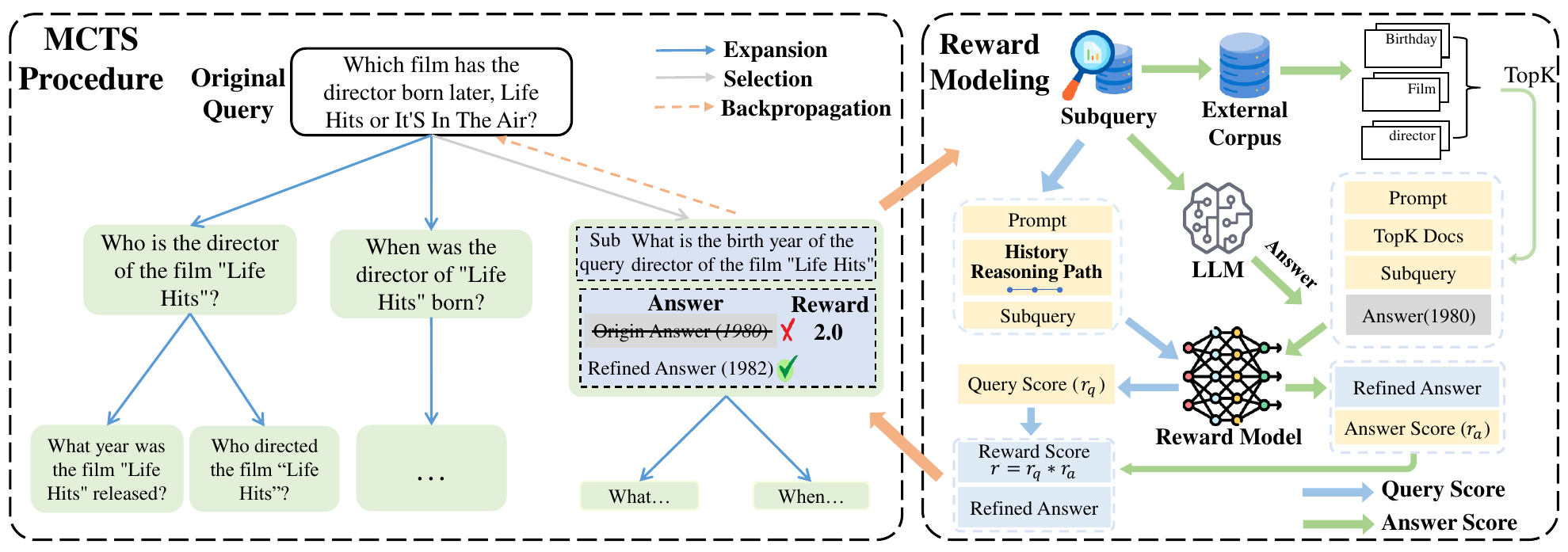}
    \caption{Overall framework of our proposed RAG-Star approach.
    } 
    \label{fig:Mix_CPT}
    \vspace{-0.3cm}
\end{figure*}

\subsection{Node Selection}
\label{sec:method-node-select}
To answer multi-hop questions, our framework will iterate the tree-based search process multiple times to gradually generate inference solutions in a step-by-step way. In our work, the solution is composed of a sequence of intermediate sub-queries and the associated answers. At each iteration, it first selects an appropriate node from the current tree for the next exploration or expansion. The selection operation is based on the node values computed through the reward modeling and backpropagation steps.

Specifically, starting from the root node $s_0$ (\ie the input question $q_0$), our RAG-Star model selects one node with the highest score {from its child nodes}, and then sequentially selects the next best child node layer-by-layer along the tree until reaching a leaf node, \ie the terminal state indicating the final answer. To better balance exploration and exploitation, we use the UCT algorithm~\citep{uct} to calculate the score of each node according to its number of visits $N(s)$ and expected reward $V(s)$ in Eq.~\ref{eq:uct}. 



\subsection{Plan Expansion}
\label{sec:method-plan-expand}
After selecting the current node, it  expands the search tree by repeatively sampling multiple child nodes as \emph{plan} based on the policy model $\pi_\theta$. Specially, the expansion process involves two steps, \ie sub-query generation and answer deduction.

\paratitle{Sub-query Planning.} To generate the next sub-query as plan, our approach first builds the context information by concatenating states from the root node to the current selected node, and then instructs the policy model to sample the next sub-query based on the context information.
Formally, given the node $s_t$, we can extract a path from the root node $s_0$ to the current node $s_t$, denoted by $\mathcal{H}=\{q_0;\langle q_1, a_1 \rangle;...;\langle q_t, a_t \rangle \}$, where $q_0$ is the original input question and each $\langle q_i, a_i \rangle$ pair denotes the planned sub-query and its answer verified by our retrieval-augmented varification (Section~\ref{sec:method-reward-modeling}).
We convert this path into the context information, and feed it to the policy model $\pi_\theta$ to generate the next sub-query $q_{t+1} = \pi_\theta(\mathcal{H})$. During inference, we employ repeated sampling to sample sub-queries by $m_q$ times to fully exploit the policy model's inherent capabilities and obtain $m_q$ new expanded sub-queries.

\paratitle{Answer Deduction.} After planning the sub-query, we further instruct the policy model to generate an answer to explore the internal knowledge of LLMs.
Specially, for each planned sub-query $q_{t+1}$, we directly feed the historical context $\mathcal{H}$ and sub-query into the policy model to generate a candidate answer by leveraging the inherent knowledge encoded in its parameters as follows:
\begin{align}\label{eq-direct-gen}
    a_{t+1}=\pi_\theta(\mathcal{H}, q_{t+1}).
\end{align}
In this process, we do not consider the external knowledge from RAG to avoid knowledge conflicts.
We aim to fully exploit the potential of the internal knowledge of LLMs without interference from external information, differing from previous retrieval-augmented work~\citep{LewisPPPKGKLYR020,Yao-arxiv-2022-ReAct} that might suffer from knowledge conflicts and interference.
After obtaining the answer, we can store each $\langle q_{t+1}, {a_{t+1}} \rangle$ pair in the corresponding node state, which will be subsequently used for reward modeling.

When completing the plan expansion process, we can obtain $m_q$ child nodes for every parent node, each of which contains a sub-query $q_{t+1}$ and its answer $a_{t+1}$.


\subsection{Reward Modeling and Backpropagation}
\label{sec:method-reward-modeling}

Traditional MCTS methods require to perform expensive rollout from the current node until the task ends to evaluate the expanded nodes. In our work, following previous work on process-supervised reward modeling~\citep{setlur2024rewarding,LightmanKBEBLLS24}, 
we propose \emph{retrieval-augmented verification and refinement} by using external knowledge to verify the consistency between the model output and retrieved information. 
Specifially, we employ reward models to assign an estimated reward $r$ to the expanded node, which effectively quantifies the effectiveness of the policy model in successfully answering the input question if continually reasoning from the current node. Next, we introduce the involved two kinds of reward scores, namely \emph{answer-aware reward} and \emph{query-aware reward}. 

\paratitle{Answer-aware Reward}. We first introduce the answer-aware reward in the verification process. First, given a sub-query $q_{t+1}$, we follow existing methods~\citep{rag} to retrieve top-$K$ documents $\mathcal{D}_{t+1}=\{d_{t+1,k}\}_{k=1}^{K}$ from the external corpus. Based on the retrieved documents, we then employ the reward model to assign an \emph{answer-aware reward} $r_a$ to the currently generated answer $a_{t+1}$ from the internal knowledge of LLMs. Specifically, there are overall three cases for the knowledge consistency between $a_{t+1}$ and $\mathcal{D}_{t+1}$ with different rewards:
\begin{equation}
r_a=\left\{
\begin{aligned}
1, \quad &\text{if}~a_{t+1}~\text{cannot be verified by}~\mathcal{D}_{t+1}& \\
2, \quad &\text{if}~a_{t+1}~\text{is in conflict with}~\mathcal{D}_{t+1}& \\
3, \quad &\text{if}~a_{t+1}~\text{is aligned with}~\mathcal{D}_{t+1}&
\end{aligned}
\right. \nonumber
\end{equation}
Note that in the second case (\ie $a_{t+1}$ is in conflict with $\mathcal{D}_{t+1}$), we assign a moderate score $2$ to the answer because we will refine $a_{t+1}$ with a new potential answer $\tilde{a}_{t+1}$ from the external knowledge $\mathcal{D}_{t+1}$ to support the policy model to continually reason from the current node. However, if the answer $a_{t+1}$ cannot be verified by the external knowledge, we will assign the lowest score $1$ to the answer, avoiding the policy model from exploring the potentially risky solution space.

\paratitle{Query-aware Reward}. In addition to evaluating the consistency of the generated answer with external knowledge, we employ the reward model to provide a \emph{query-aware reward} $r_q$ for measuring the plausibility of the planned sub-query $q_{t+1}$ based on the historical context information from the root node to current node, \ie $\mathcal{H}=\{q_0;\langle q_1, a_1 \rangle;...;\langle q_t, a_t \rangle \}$. If the sub-query evaluated by the reward model is logically inconsistent with the history plan, the score $r_q$ is set to 0; otherwise, it is set to 1. Therefore, the final reward $r$ for the expanded node $s_{t+1}$ is computed as $r = r_a \cdot r_q$. This step aims to prevent the policy model from continuing to reason along illogical sub-queries. 

After obtaining the final reward for the newly expanded node, we backpropagate the reward to update the value of nodes from the root node $s_0$ to the current node $s_{t+1}$. For each node $s_0,s_1,...,s_{t+1}$ in the path, its number of visits $N(s)$ and the value $V(s)$ will be updated according to Eq.~\ref{eq:back}. These updated values are used in the UCT algorithm in Eq.~\ref{eq:uct} to guide the node selection at the next iteration. 


\subsection{Reward Model Training}


In the reward modeling process, the capacity of the reward model critically influences the search process and ultimate answer accuracy. 
However, utilizing close-source model API or very large LLMs incurs substantial computational costs for deployment. Hence, we adopt a knowledge distillation technique to transfer capabilities from an advanced LLM, which usually has more parameters, to a relatively smaller model. This involves two phases: data synthesis and instruction fine-tuning.

During data synthesis, we mix up training sets from our evaluation datasets to maintain diversity. First, we adopt in-context learning to instruct the policy model to generate a CoT format solution and then break down into multiple sub-steps, each incorporating the input question, accumulated reasoning paths, and a sub-query specific to the current step. To further ensure diversity, only one random step from each sample is selected for subsequent instruction data creation. We then employ a more advanced LLM (\ie GPT-4o-mini) combined with a retrieval system to evaluate the sub-query and its answer for each step (Section~\ref{sec:method-reward-modeling}), and filter the output that fails to meet the format criteria. Finally, we compile a dataset of intermediate steps and their query and answer rewards from an advanced LLM. In the instruction fine-tuning phase, we utilize the synthetic samples to fine-tune a smaller LLM (\ie Llama-3.1-8B-Instruct), thereby enhancing its capabilities in reward modeling.

%% file: sec-experiment.tex
\section{Experiments}
\label{experiment}


\subsection{Experimental Setup}
\label{sec:app_experiment}

\paragraph{Datasets and Evaluation Metrics.}
We select four typical complex multi-hop question-answering datasets, \ie HotpotQA~\citep{hotpotqa}, 2WikiMultihopQA~\citep{2wiki}, MusiQue~\citep{musique}, and StrategyQA~\citep{strategyqa}. For evaluation metrics, we use Exact Match (EM), F1 score, and Cover Exact Match (Cover EM), where Cover EM measures whether the ground truth answer is covered in the generated answer. 
We randomly select 100 samples from the whole validation sets of each dataset as our final test set for all baselines and our method.

\begin{table*}[t]
\centering
\small
\begin{tabular}{lcccccccccccc}
\toprule
\multicolumn{1}{c}{\multirow{2.5}{*}{\textbf{Method}}} & \multicolumn{3}{c}{\textbf{HotpotQA}} & \multicolumn{3}{c}{\textbf{2WikiMultihopQA}} & \multicolumn{3}{c}{\textbf{MusiQue}} & \multicolumn{3}{c}{\textbf{StrategyQA}} \\
\cmidrule(r){2-4} \cmidrule(r){5-7} \cmidrule(r){8-10} \cmidrule(r){11-13}
\multicolumn{1}{c}{} & \textbf{EM} & \textbf{CEM} & \textbf{F1} & \textbf{EM} & \textbf{CEM} & \textbf{F1} & \textbf{EM} & \textbf{CEM} & \textbf{F1} & \textbf{EM} & \textbf{CEM} & \textbf{F1} \\
\midrule
Llama-3.1-8B-Instruct & 14.0 & 25.0 & 26.0 & 9.0 & 29.0 & 21.9 & 2.0 & 3.0 & 3.9 & 63.0 & 65.0 & 63.0 \\
\cmidrule{2-13}
\multicolumn{1}{l}{+ Chain-of-Tought} & 20.0 & 38.0 & 26.3 & 4.0 & 32.0 & 7.1 & 4.0 & 16.0 & 6.6 & 55.0 & \underline{69.0} & 55.0 \\
\multicolumn{1}{l}{+ Standard RAG} &  40.0 & \underline{48.0} & 52.8 & 17.0 & 23.0 & 26.1 & 11.0 & 11.0 & 15.5 & 63.0 & 64.0 & 63.0  \\
\multicolumn{1}{l}{+ Iterative RAG} &  26.0 & 31.0 & 36.9 & 22.0 & 23.0 & 26.0 & 7.0 & 11.0 & 15.9 & 61.0 & 63.0 & 61.0  \\
\multicolumn{1}{l}{+ Generate-then-Retrieve} &  34.0 & 44.0 & 49.4 & 21.0 & 30.0 & 26.6 & 13.0 & 17.0 & 19.4 & 63.0 & 67.0 & 63.0 \\
\multicolumn{1}{l}{+ Judge-then-Retrieve} &  39.0 & \underline{48.0} & 53.9 & 18.0 & 26.0 & 26.8 & 10.0 & 10.0 & 16.0 & 58.0 & 63.0 & 58.0 \\
\cmidrule{2-13}
\multicolumn{1}{l}{+ RAG-Star w Llama RM} & \underline{42.0} & 44.0 & \underline{54.4} & \underline{34.0} & \underline{38.0} & \underline{42.0} & \underline{13.0} & \underline{18.0} & \underline{22.2} & \textbf{71.0} & \textbf{72.0} & \textbf{71.0} \\
\multicolumn{1}{l}{+ RAG-Star w GPT RM} & \textbf{46.0} & \textbf{49.0} & \textbf{60.0} & \textbf{38.0} & \textbf{43.0} & \textbf{46.8} & \textbf{22.2} & \textbf{27.0} & \textbf{30.7} & \underline{67.6} & \underline{69.0} & \underline{67.6} \\
\midrule
GPT-4o & 43.0 & 47.0 & 56.7 & 36.0 & 42.0 & 45.7 & 13.0 & 19.0 & 24.3 & \underline{70.0} & 73.0 & \underline{70.0} \\
\cmidrule{2-13}
\multicolumn{1}{l}{+ Chain-of-Tought} & 36.0 & 49.0 & 56.8 & 38.0 & 55.0 & 53.9 & 20.0 & 27.0 & 29.6 & 37.0 & 79.0 & 37.0 \\
\multicolumn{1}{l}{+ Standard RAG} & \underline{47.0} & \underline{57.0} & 63.7 & 25.0 & 26.0 & 31.2 & 14.0 & 18.0 & 20.6 & 45.0 & 62.0 & 45.0 \\
\multicolumn{1}{l}{+ Iterative RAG} & \underline{47.0} & \textbf{59.0} & 63.3 & 19.0 & 24.0 & 26.3 & 15.0 & 26.0 & 25.5 & 32.0 & 74.0 & 32.0 \\
\multicolumn{1}{l}{+ Generate-then-Retrieve} & 44.0 & \underline{57.0} & 62.0 & 29.0 & 36.0 & 37.5 & 23.0 & 28.0 & 31.0 & 50.0 & 68.0 & 50.0 \\
\multicolumn{1}{l}{+ Judge-then-Retrieve} & 44.0 & 50.0 & 58.6 & 28.0 & 29.0 & 32.2 & 14.0 & 16.0 & 22.8 & \textbf{72.0} & 74.0 & \textbf{72.0} \\
\cmidrule{2-13}
\multicolumn{1}{l}{+ RAG-Star w Llama RM} & \textbf{48.0} & 54.0 & \underline{66.3} & \underline{47.0} & \textbf{68.0} & \textbf{62.8} & \underline{25.0} & \underline{36.0} & \underline{39.0} & {61.0} & \textbf{86.0} & 61.0 \\
\multicolumn{1}{l}{+ RAG-Star w GPT RM} & \textbf{48.0} & \underline{57.0} & \textbf{68.6} & \textbf{48.0} & \underline{63.0} & \underline{61.7} & \textbf{29.0} & \textbf{40.0} & \textbf{43.5} & 60.0 & \underline{81.0} & 60.0 \\
\bottomrule
\end{tabular}
\caption{Evaluation results on four representative multi-hop question answering tasks. ``RM'' is short for reward model. {The \textbf{bold} and \underline{underline} fonts denote the best and second best results in each dataset, respectively.}}
\label{tab:main-result}
\end{table*}

\paragraph{Baselines.}

We compare \textbf{RAG-Star} to the following two types of baselines based on GPT-4o and Llama-3.1-8B-Instruct:


\textbullet~\textbf{Vanilla prompting methods} including direct prompting, Chain-of-Thought (CoT), and standard RAG. Direct prompting instructs the model to directly generate answers and CoT incorporates intermediate reasoning steps, which are all based on the inherent knowledge of LLMs. Standard RAG first retrieves documents from Wikipedia based on DPR~\citep{dpr} as prompts and then generates the final answers.


\textbullet~\textbf{Improved RAG methods} including Iterative RAG~\citep{searchain}, Judge-then-retrieve~\citep{self-rag}, and Generate-then-retrieve~\citep{query2doc}. We reimplement all of these baselines in our experiments. Iterative RAG iteratively decomposes the input question into sub-queries for retrieval and generation, ultimately ensembling all intermediate answers; Judge-then-retrieve first decides whether the retrieval is needed, autonomously deciding to utilize either internal or external knowledge to aid in generation; Generate-then-retrieve first generates an initial answer used for retrieving more documents relevant to the question and then generates the final answer based on documents.


\paragraph{Implementation Details.}
We use a closed-source model (GPT-4o) and an open-source model (Llama-3.1-8B-Instruct) as our policy models to measure the performance of the RAG-Star framework. For the reward models, we use GPT-4o-mini and a fine-tuned Llama-3.1-8B-Instruct.
For HotpotQA, we only use the abstract of articles in Wikipedia 2017 dump as the retrieval corpus following \citet{hotpotqa}, while for other datasets, we use the whole articles in Wikipedia 2018 dump~\citep{dpr}. Moreover, for the retrieval model, we use FAISS for index building and BGE-large-en-v1.5~\citep{bge_embedding} for dense passage retrieval.
For all retrieval-based baselines, we retrieve top-$5$ documents and employ greedy search for decoding with a temperature of $0$. 
For RAG-Star, we set the maximum number of simulations to $50$ and a maximum of $6$ layers. 
In UCT algorithm, the weight $w$ to control the exploration and exploitation is set to $0.2$. 
We also retrieve top-$5$ documents and sample three sub-queries at a time ($m_q = 3$) with temperature $1.0$ and top-$p$ sampling where $p = 1.0$.
For answer generation, we sample an answer using a temperature of $0.9$ and top-$p$ sampling set to $1.0$.


\subsection{Main Results}

Table~\ref{tab:main-result} shows the results of RAG-Star and other baselines across four representative multi-hop question answering datasets.


Firstly, it can be observed that relatively smaller models (\eg Llama-3.1-8B-Instruct) show limited performance on these knowledge-intensive reasoning tasks, achieving below 10\% across three metrics in MusiQue. Although the Chain-of-Thought technique can slightly improve the answer recall (\eg Cover EM scores of Llama-3.1-8B-Instruct and GPT-4o in MusiQue increase from 3.0\% and 19.0\% to 16.0\% and 27.0\%, respectively), the model is prone to generating substantial irrelevant information in the output, decreasing the overall performance (\eg F1 score of Llama-3.1-8B-Instruct drops from 21.9\% to 7.1\% on 2WikiMultihopQA).

Secondly, based on the standard RAG, GPT-4o achieves substantial improvement in HotpotQA (\eg Cover EM increases from 47.0\% to 57.0\%) but exhibits a large decline in StrategyQA (\eg Cover EM from 73.0\% to 62.0\%), suggesting a potential conflict between external sources and internal knowledge of LLMs. We speculate the reason might be that using the retrieved information directly as input incorporates some noises and makes the LLM lost in the useful information. Therefore, by controlling the utilization of internal and external knowledge, Judge-then-Retrieve can significantly alleviate this issue (\eg Cover EM from 62.0\% to 74.0\% in StrategyQA). However, these approaches still present limited or even negative improvements in complex tasks (\eg Cover EM from 19.0\% to 16.0\% in MusiQue), necessitating effective methods to consolidate external and internal knowledge.


{Finally, our approach outperforms all baselines across most metrics in four datasets. RAG-Star introduces a ``System 2''-like slow and deliberative thinking process and employs RAG to verify and guide the multi-step reasoning process. By employing retrieval-augmented verification, the reward model can effectively encourage the model towards plausible sub-query nodes or avert from potential risky nodes. For example, equipped with our RAG-Star framework, Llama-3.1-8B-Instruct achieves higher scores in two challenging reasoning datasets, \ie 2WikiMultihopQA and MusiQue, significantly beyond all baseline methods.
}

\subsection{Further Analysis}

We report further analysis in StrategyQA with randomly selected 100 samples -- we have similar findings in other datasets.

\begin{table}[t]
\centering
\small
\begin{tabular}{l  c c  c c }
    \toprule
    \multirow{2.5}{*}{\textbf{Method}} & \multicolumn{2}{c}{\textbf{GPT-4o}} & \multicolumn{2}{c}{\textbf{Llama3.1-8B}} \\
    \cmidrule(r){2-3}\cmidrule(r){4-5}
    & \textbf{CEM} & \textbf{F1} & \textbf{CEM} & \textbf{F1} \\
    \midrule
    RAG-Star~(Ours) & 84.0 & 68.3 & 75.0 & 73.3 \\
    \midrule
    w/o Query Score  & 82.0 & 68.0 & 71.0 & 69.0 \\
    w/o Answer Score & 80.0 & 66.3 & 66.0 & 65.3\\
    w/o Retrieval & 78.0 & 67.3 & 67.0 & 66.0\\
    w/o Refine & 77.0 & 68.2 & 70.0 & 68.1\\
    \bottomrule
\end{tabular}
\caption{Ablation study in StrategyQA.}
\label{tab:ablation}
\end{table}

\paragraph{Ablation Study.} To validate the effectiveness of our proposed framework, we conduct an ablation analysis of its key design elements. 
We design four variants: (1) \emph{w/o Retrieval} removes the retrieved documents in reward modeling; (2) \emph{w/o Refine} does not refine the conflict answer with retrieved documents in reward modeling; (3) \emph{w/o Query Reward} removes the query-aware reward $r_q$ for scoring; and (4) \emph{w/o Answer Reward} removes the answer-aware reward $r_a$ for scoring.
We show the results in Table~\ref{tab:ablation}. It is clear that all the variants perform worse than the original method, indicating the effectiveness of each component in our framework. Specifically, the performance of \emph{w/o Retrieval} drops significantly for Llama-3.1-8B, indicating that using external knowledge for verification can be highly beneficial for the inherent reasoning of LLMs. Similarly, \emph{w/o Refine} leads to a decline in model performance, which highlights the importance of repurposing external sources for correcting the errors in the model's reasoning process. Moreover,  both \emph{w/o Query Reward} and \emph{w/o Answer Reward} variants lead to a substantial performance decline, which suggests that the consistency and logical plausibility of intermediate sub-queries and answers are both critical for the model to plan the correct path towards the final answer. 


\begin{figure}[t]
    \centering
    \small
    \includegraphics[width=0.48\textwidth]{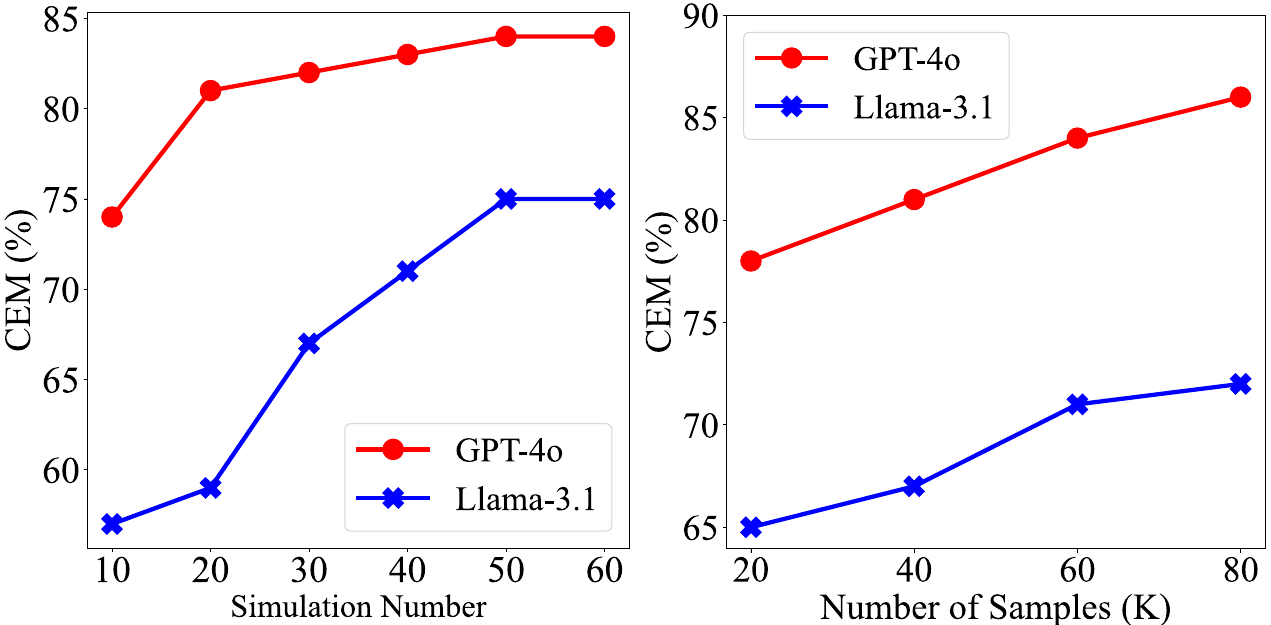}
    \caption{Cover EM performance on the StrategyQA \emph{w.r.t.} the number of simulations (\textbf{Left}) or the number of training data (\textbf{Right}).
    } 
    \label{fig:simulation_acc}
    \vspace{-0.3cm}
\end{figure}

\begin{figure*}[t]
    \centering
    \includegraphics[width=\textwidth]{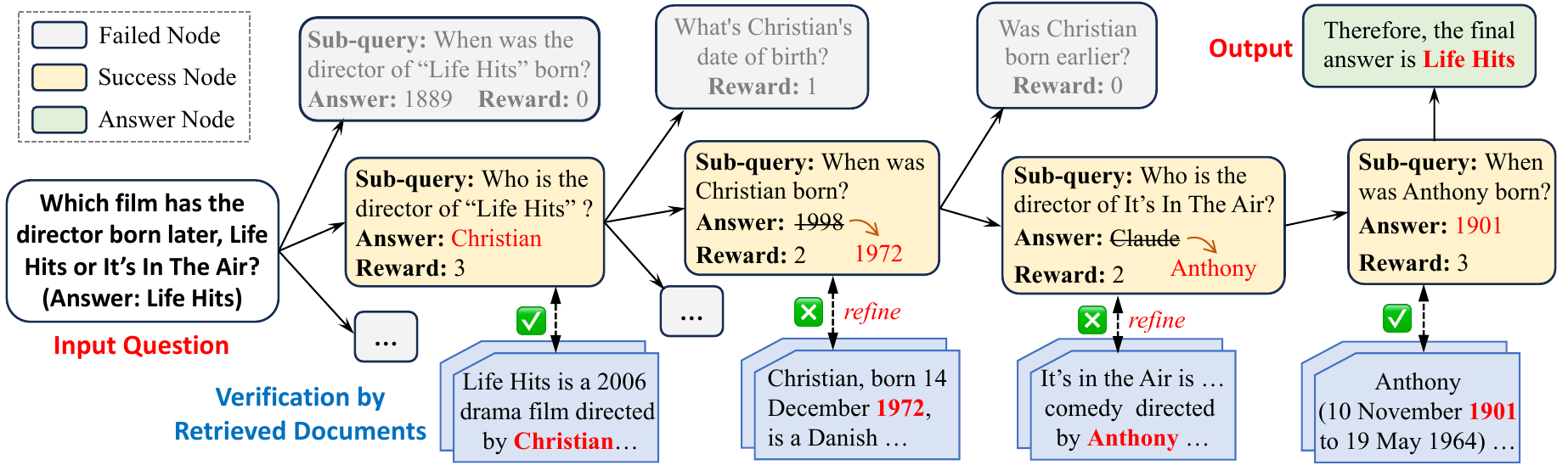}
    \caption{A qualitative example showing the deliberative reasoning process of RAG-Star in 2WikiMultihopQA. 
    } 
    \label{fig:case}
\end{figure*}

\paragraph{Effect of Simulation Scaling.}
Typically, scaling the simulation iterations will lead to a higher level of task-solving capability. To explore the relationship between simulation scaling and the final performance of RAG-Star, we test our model under different maximum simulation iterations.
Specifically, we vary the maximum simulation rounds in a set $\{ 10, 20, 30, 40, 50, 60 \}$, and evaluate Llama and GPT-4o in StrategyQA with GPT-4o-mini as the reward model. The results are presented in the Figure~\ref{fig:simulation_acc}. We can see that as the maximum number of simulation increases, the model's performance gradually improves, although the average time consumed also rises to some extent. This highlights that scaling the test-time computation can further promote more thorough exploration and exploitation by the policy model within the search space. However, as the number of simulations further increases, the performance of the policy model tends to be saturated. Due to the limitation of inherent knowledge, the policy model cannot benefit a lot from conducting more simulations. 



\paragraph{Effect of Reward Model.}
In our framework, the reward model is used to assess the logical plausibility of the sub-query and the consistency between the output answer and external sources. In this part, we aim to explore how to train open-source reward models~(\ie Llama-3.1-8B-Instruct) to achieve performance comparable to closed-source LLMs~(\ie GPT-4o-mini) by varying amounts of training data from 20K to 80K.
Specifically, we employ different amounts of training data to fine-tune Llama-3.1-8B-Instruct and use the fine-tuned model to evaluate the sub-query and its answer. As shown in Figure~\ref{fig:simulation_acc}, we can see that as the amount of training data increases, the reward model can achieve more accurate verification quality, significantly benefiting the planning and reasoning of the policy model. However, the performance gains tend to saturate at later stages, necessitating instruction tuning data with higher diversity and quality. 


\subsection{Case Study}
To facilitate understanding of the entire workflow of our proposed RAG-Star, we present a qualitative analysis in 2WikiMultihopQA. Throughout the search process, the LLM initializes the input question as root node and conducts multiple simulations, eventually reaching the terminal leaf node, which can be vividly represented as a tree. 
As shown in Figure~\ref{fig:case}, after selecting the first query (\ie \emph{Who is the director of ``Life Hits''?}), the model expands multiple children nodes by repeated sampling. At the next iteration, the model refines the generated answer (\ie \emph{1998}) for the sub-query (``\emph{When was Christian born?}'') based on retrieved documents and the reward model returns an overall score of $2$. By iterating the multi-step reasoning and retrieval-augmented verification processes for several rounds, the model outputs the final answer (\ie \emph{Life Hits}). In the task-solving process, the policy model generates an answer to the current sub-query based on its internal knowledge, which might be erroneous due to the limited pre-training corpus in time or the memorization mistakes. Therefore, the external knowledge can be beneficial to validate the correctness of inherent knowledge of LLMs, effectively guiding the model to plan a reasonable path.


%% file: sec-conclusion.tex
\section{Conclusion}
\label{conclusion}

In this work, we proposed RAG-Star, a novel RAG approach for leveraging external retrieval technique to enhance the multi-step reasoning capabilities of LLMs. RAG-Star employed Monte Carlo Tree Search to search intermediate sub-queries and corresponding answers. Moreover, RAG-Star introduced retrieval-augmented verification to evaluate the plausibility and consistency of the planned sub-queries and answers based on a query-aware and an answer-aware reward. At each iteration, RAG-Star conducted node selection, plan expansion, reward modeling, and reward backpropagation sequentially to consolidate the internal knowledge of LLMs and external knowledge from RAG. Extensive experiments on several datasets showed
that our proposed RAG-Star outperforms the traditional RAG and reasoning methods.

%% file: sec-limitation.tex
\section*{Limitations} \label{limitations}

Despite the great efforts that we have made, the experimental analysis is still limited due to the massive computational cost of tree-based search approaches. We will investigate into more types of complex reasoning tasks and datasets. In our model, we only leverage Monte Carlo Tree Search to conduct our deleberative reasoning process. we may consider investigate more kinds of search algorithms to verify the generalization and robustness of our proposed framework. Moreover, the performance of our model is affected by the feedback quality provided by the reward model. Therefore, a well-trained and performant reward model is important for guiding the reasoning process. We will consider other fine-tuning strategies and more LLMs in reward modeling.

%% file: acl2024.bbl
\begin{thebibliography}{41}
\expandafter\ifx\csname natexlab\endcsname\relax\def\natexlab#1{#1}\fi

\bibitem[{Asai et~al.(2024)Asai, Wu, Wang, Sil, and Hajishirzi}]{self-rag}
Akari Asai, Zeqiu Wu, Yizhong Wang, Avirup Sil, and Hannaneh Hajishirzi. 2024.
\newblock Self-rag: Learning to retrieve, generate, and critique through self-reflection.
\newblock In \emph{The Twelfth International Conference on Learning Representations, {ICLR} 2024, Vienna, Austria, May 7-11, 2024}. OpenReview.net.

\bibitem[{Borgeaud et~al.(2022)Borgeaud, Mensch, Hoffmann, Cai, Rutherford, Millican, Van Den~Driessche, Lespiau, Damoc, Clark et~al.}]{borgeaud2022improving}
Sebastian Borgeaud, Arthur Mensch, Jordan Hoffmann, Trevor Cai, Eliza Rutherford, Katie Millican, George~Bm Van Den~Driessche, Jean-Baptiste Lespiau, Bogdan Damoc, Aidan Clark, et~al. 2022.
\newblock Improving language models by retrieving from trillions of tokens.
\newblock In \emph{International conference on machine learning}, pages 2206--2240. PMLR.

\bibitem[{Brown et~al.(2024)Brown, Juravsky, Ehrlich, Clark, Le, R{\'{e}}, and Mirhoseini}]{abs-2407-21787}
Bradley C.~A. Brown, Jordan Juravsky, Ryan~Saul Ehrlich, Ronald Clark, Quoc~V. Le, Christopher R{\'{e}}, and Azalia Mirhoseini. 2024.
\newblock \href {https://doi.org/10.48550/ARXIV.2407.21787} {Large language monkeys: Scaling inference compute with repeated sampling}.
\newblock \emph{CoRR}, abs/2407.21787.

\bibitem[{Brown et~al.(2020)Brown, Mann, Ryder, Subbiah, Kaplan, Dhariwal, Neelakantan, Shyam, Sastry, Askell, Agarwal, Herbert{-}Voss, Krueger, Henighan, Child, Ramesh, Ziegler, Wu, Winter, Hesse, Chen, Sigler, Litwin, Gray, Chess, Clark, Berner, McCandlish, Radford, Sutskever, and Amodei}]{gpt3}
Tom~B. Brown, Benjamin Mann, Nick Ryder, Melanie Subbiah, Jared Kaplan, Prafulla Dhariwal, Arvind Neelakantan, Pranav Shyam, Girish Sastry, Amanda Askell, Sandhini Agarwal, Ariel Herbert{-}Voss, Gretchen Krueger, Tom Henighan, Rewon Child, Aditya Ramesh, Daniel~M. Ziegler, Jeffrey Wu, Clemens Winter, Christopher Hesse, Mark Chen, Eric Sigler, Mateusz Litwin, Scott Gray, Benjamin Chess, Jack Clark, Christopher Berner, Sam McCandlish, Alec Radford, Ilya Sutskever, and Dario Amodei. 2020.
\newblock Language models are few-shot learners.
\newblock In \emph{Advances in Neural Information Processing Systems 33: Annual Conference on Neural Information Processing Systems 2020, NeurIPS 2020, December 6-12, 2020, virtual}.

\bibitem[{Chen et~al.(2019)Chen, Lin, and Durrett}]{multi-hop}
Jifan Chen, Shih{-}Ting Lin, and Greg Durrett. 2019.
\newblock \href {http://arxiv.org/abs/1910.02610} {Multi-hop question answering via reasoning chains}.
\newblock \emph{CoRR}, abs/1910.02610.

\bibitem[{Gao et~al.(2023)Gao, Xiong, Gao, Jia, Pan, Bi, Dai, Sun, and Wang}]{gao2023retrieval}
Yunfan Gao, Yun Xiong, Xinyu Gao, Kangxiang Jia, Jinliu Pan, Yuxi Bi, Yi~Dai, Jiawei Sun, and Haofen Wang. 2023.
\newblock Retrieval-augmented generation for large language models: A survey.
\newblock \emph{arXiv preprint arXiv:2312.10997}.

\bibitem[{Geva et~al.(2021)Geva, Khashabi, Segal, Khot, Roth, and Berant}]{strategyqa}
Mor Geva, Daniel Khashabi, Elad Segal, Tushar Khot, Dan Roth, and Jonathan Berant. 2021.
\newblock Did aristotle use a laptop? {A} question answering benchmark with implicit reasoning strategies.
\newblock \emph{Trans. Assoc. Comput. Linguistics}, 9:346--361.

\bibitem[{Guu et~al.(2020)Guu, Lee, Tung, Pasupat, and Chang}]{realm}
Kelvin Guu, Kenton Lee, Zora Tung, Panupong Pasupat, and Mingwei Chang. 2020.
\newblock Retrieval augmented language model pre-training.
\newblock In \emph{International Conference on Machine Learning}, pages 3929--3938. PMLR.

\bibitem[{Hart et~al.(1968)Hart, Nilsson, and Raphael}]{HartNR68}
Peter~E. Hart, Nils~J. Nilsson, and Bertram Raphael. 1968.
\newblock A formal basis for the heuristic determination of minimum cost paths.
\newblock \emph{{IEEE} Trans. Syst. Sci. Cybern.}, 4(2):100--107.

\bibitem[{Ho et~al.(2020)Ho, Nguyen, Sugawara, and Aizawa}]{2wiki}
Xanh Ho, Anh{-}Khoa~Duong Nguyen, Saku Sugawara, and Akiko Aizawa. 2020.
\newblock Constructing {A} multi-hop {QA} dataset for comprehensive evaluation of reasoning steps.
\newblock In \emph{Proceedings of the 28th International Conference on Computational Linguistics, {COLING} 2020, Barcelona, Spain (Online), December 8-13, 2020}, pages 6609--6625. International Committee on Computational Linguistics.

\bibitem[{Huang and Chang(2023)}]{reasoning_survey}
Jie Huang and Kevin~Chen{-}Chuan Chang. 2023.
\newblock Towards reasoning in large language models: {A} survey.
\newblock In \emph{Findings of the Association for Computational Linguistics: {ACL} 2023, Toronto, Canada, July 9-14, 2023}, pages 1049--1065. Association for Computational Linguistics.

\bibitem[{Huang et~al.(2024)Huang, Wang, Xia, Li, Zou, Xu, Fan, Ye, Chern, Ye, Zhang, Yang, Wu, Wang, Sun, Xiao, Li, Zhou, Chern, Qin, Ma, Su, Liu, Zheng, Zhang, Lin, Qiao, and Liu}]{abs-2406-12753}
Zhen Huang, Zengzhi Wang, Shijie Xia, Xuefeng Li, Haoyang Zou, Ruijie Xu, Run{-}Ze Fan, Lyumanshan Ye, Ethan Chern, Yixin Ye, Yikai Zhang, Yuqing Yang, Ting Wu, Binjie Wang, Shichao Sun, Yang Xiao, Yiyuan Li, Fan Zhou, Steffi Chern, Yiwei Qin, Yan Ma, Jiadi Su, Yixiu Liu, Yuxiang Zheng, Shaoting Zhang, Dahua Lin, Yu~Qiao, and Pengfei Liu. 2024.
\newblock Olympicarena: Benchmarking multi-discipline cognitive reasoning for superintelligent {AI}.
\newblock \emph{CoRR}, abs/2406.12753.

\bibitem[{Kahneman(2011)}]{kahneman2011thinking}
Daniel Kahneman. 2011.
\newblock Thinking, fast and slow.
\newblock \emph{Farrar, Straus and Giroux}.

\bibitem[{Karpukhin et~al.(2020)Karpukhin, O{\u{g}}uz, Min, Lewis, Wu, Edunov, Chen, and Yih}]{dpr}
Vladimir Karpukhin, Barlas O{\u{g}}uz, Sewon Min, Patrick Lewis, Ledell Wu, Sergey Edunov, Danqi Chen, and Wen-tau Yih. 2020.
\newblock Dense passage retrieval for open-domain question answering.
\newblock \emph{arXiv preprint arXiv:2004.04906}.

\bibitem[{Kocsis and Szepesv{\'{a}}ri(2006)}]{uct}
Levente Kocsis and Csaba Szepesv{\'{a}}ri. 2006.
\newblock Bandit based monte-carlo planning.
\newblock In \emph{Machine Learning: {ECML} 2006, 17th European Conference on Machine Learning, Berlin, Germany, September 18-22, 2006, Proceedings}, volume 4212 of \emph{Lecture Notes in Computer Science}, pages 282--293. Springer.

\bibitem[{Lewis et~al.(2020{\natexlab{a}})Lewis, Perez, Piktus, Petroni, Karpukhin, Goyal, K{\"u}ttler, Lewis, Yih, Rockt{\"a}schel et~al.}]{rag}
Patrick Lewis, Ethan Perez, Aleksandra Piktus, Fabio Petroni, Vladimir Karpukhin, Naman Goyal, Heinrich K{\"u}ttler, Mike Lewis, Wen-tau Yih, Tim Rockt{\"a}schel, et~al. 2020{\natexlab{a}}.
\newblock Retrieval-augmented generation for knowledge-intensive nlp tasks.
\newblock \emph{Advances in Neural Information Processing Systems}, 33:9459--9474.

\bibitem[{Lewis et~al.(2020{\natexlab{b}})Lewis, Perez, Piktus, Petroni, Karpukhin, Goyal, K{\"{u}}ttler, Lewis, Yih, Rockt{\"{a}}schel, Riedel, and Kiela}]{LewisPPPKGKLYR020}
Patrick S.~H. Lewis, Ethan Perez, Aleksandra Piktus, Fabio Petroni, Vladimir Karpukhin, Naman Goyal, Heinrich K{\"{u}}ttler, Mike Lewis, Wen{-}tau Yih, Tim Rockt{\"{a}}schel, Sebastian Riedel, and Douwe Kiela. 2020{\natexlab{b}}.
\newblock Retrieval-augmented generation for knowledge-intensive {NLP} tasks.
\newblock In \emph{Advances in Neural Information Processing Systems 33: Annual Conference on Neural Information Processing Systems 2020, NeurIPS 2020, December 6-12, 2020, virtual}.

\bibitem[{Lightman et~al.(2024)Lightman, Kosaraju, Burda, Edwards, Baker, Lee, Leike, Schulman, Sutskever, and Cobbe}]{LightmanKBEBLLS24}
Hunter Lightman, Vineet Kosaraju, Yuri Burda, Harrison Edwards, Bowen Baker, Teddy Lee, Jan Leike, John Schulman, Ilya Sutskever, and Karl Cobbe. 2024.
\newblock Let's verify step by step.
\newblock In \emph{The Twelfth International Conference on Learning Representations, {ICLR} 2024, Vienna, Austria, May 7-11, 2024}. OpenReview.net.

\bibitem[{Lyu et~al.(2023)Lyu, Havaldar, Stein, Zhang, Rao, Wong, Apidianaki, and Callison{-}Burch}]{LyuHSZRWAC23}
Qing Lyu, Shreya Havaldar, Adam Stein, Li~Zhang, Delip Rao, Eric Wong, Marianna Apidianaki, and Chris Callison{-}Burch. 2023.
\newblock Faithful chain-of-thought reasoning.
\newblock In \emph{Proceedings of the 13th International Joint Conference on Natural Language Processing and the 3rd Conference of the Asia-Pacific Chapter of the Association for Computational Linguistics, {IJCNLP} 2023 -Volume 1: Long Papers, Nusa Dua, Bali, November 1 - 4, 2023}, pages 305--329. Association for Computational Linguistics.

\bibitem[{OpenAI(2023)}]{OpenAI-OpenAI-2023-GPT-4}
OpenAI. 2023.
\newblock Gpt-4 technical report.
\newblock \emph{OpenAI Blog}.

\bibitem[{Setlur et~al.(2024)Setlur, Nagpal, Fisch, Geng, Eisenstein, Agarwal, Agarwal, Berant, and Kumar}]{setlur2024rewarding}
Amrith Setlur, Chirag Nagpal, Adam Fisch, Xinyang Geng, Jacob Eisenstein, Rishabh Agarwal, Alekh Agarwal, Jonathan Berant, and Aviral Kumar. 2024.
\newblock Rewarding progress: Scaling automated process verifiers for llm reasoning.
\newblock \emph{arXiv preprint arXiv:2410.08146}.

\bibitem[{Silver et~al.(2016)Silver, Huang, Maddison, Guez, Sifre, van~den Driessche, Schrittwieser, Antonoglou, Panneershelvam, Lanctot, Dieleman, Grewe, Nham, Kalchbrenner, Sutskever, Lillicrap, Leach, Kavukcuoglu, Graepel, and Hassabis}]{alphago}
David Silver, Aja Huang, Chris~J. Maddison, Arthur Guez, Laurent Sifre, George van~den Driessche, Julian Schrittwieser, Ioannis Antonoglou, Vedavyas Panneershelvam, Marc Lanctot, Sander Dieleman, Dominik Grewe, John Nham, Nal Kalchbrenner, Ilya Sutskever, Timothy~P. Lillicrap, Madeleine Leach, Koray Kavukcuoglu, Thore Graepel, and Demis Hassabis. 2016.
\newblock Mastering the game of go with deep neural networks and tree search.
\newblock \emph{Nat.}, 529(7587):484--489.

\bibitem[{Silver et~al.(2017)Silver, Hubert, Schrittwieser, Antonoglou, Lai, Guez, Lanctot, Sifre, Kumaran, Graepel, Lillicrap, Simonyan, and Hassabis}]{alphazero}
David Silver, Thomas Hubert, Julian Schrittwieser, Ioannis Antonoglou, Matthew Lai, Arthur Guez, Marc Lanctot, Laurent Sifre, Dharshan Kumaran, Thore Graepel, Timothy~P. Lillicrap, Karen Simonyan, and Demis Hassabis. 2017.
\newblock \href {http://arxiv.org/abs/1712.01815} {Mastering chess and shogi by self-play with a general reinforcement learning algorithm}.
\newblock \emph{CoRR}, abs/1712.01815.

\bibitem[{Snell et~al.(2024)Snell, Lee, Xu, and Kumar}]{abs-2408-03314}
Charlie Snell, Jaehoon Lee, Kelvin Xu, and Aviral Kumar. 2024.
\newblock \href {https://doi.org/10.48550/ARXIV.2408.03314} {Scaling {LLM} test-time compute optimally can be more effective than scaling model parameters}.
\newblock \emph{CoRR}, abs/2408.03314.

\bibitem[{Sutton(2019)}]{sutton2019bitter}
Richard Sutton. 2019.
\newblock The bitter lesson.
\newblock \emph{Incomplete Ideas (blog)}, 13(1):38.

\bibitem[{Suzgun et~al.(2023)Suzgun, Scales, Sch{\"{a}}rli, Gehrmann, Tay, Chung, Chowdhery, Le, Chi, Zhou, and Wei}]{bbh}
Mirac Suzgun, Nathan Scales, Nathanael Sch{\"{a}}rli, Sebastian Gehrmann, Yi~Tay, Hyung~Won Chung, Aakanksha Chowdhery, Quoc~V. Le, Ed~H. Chi, Denny Zhou, and Jason Wei. 2023.
\newblock Challenging big-bench tasks and whether chain-of-thought can solve them.
\newblock In \emph{Findings of the Association for Computational Linguistics: {ACL} 2023, Toronto, Canada, July 9-14, 2023}, pages 13003--13051. Association for Computational Linguistics.

\bibitem[{Trivedi et~al.(2022)Trivedi, Balasubramanian, Khot, and Sabharwal}]{musique}
Harsh Trivedi, Niranjan Balasubramanian, Tushar Khot, and Ashish Sabharwal. 2022.
\newblock Musique: Multihop questions via single-hop question composition.
\newblock \emph{Trans. Assoc. Comput. Linguistics}, 10:539--554.

\bibitem[{Wang et~al.(2024{\natexlab{a}})Wang, Deng, Lv, Liang, He, Yan, and An}]{abs-2406-14283}
Chaojie Wang, Yanchen Deng, Zhiyi Lv, Zeng Liang, Jujie He, Shuicheng Yan, and Bo~An. 2024{\natexlab{a}}.
\newblock \href {https://doi.org/10.48550/ARXIV.2406.14283} {Q*: Improving multi-step reasoning for llms with deliberative planning}.
\newblock \emph{CoRR}, abs/2406.14283.

\bibitem[{Wang et~al.(2024{\natexlab{b}})Wang, Wan, Sun, Chen, and Ar{\i}k}]{wang2024astute}
Fei Wang, Xingchen Wan, Ruoxi Sun, Jiefeng Chen, and Sercan~{\"O} Ar{\i}k. 2024{\natexlab{b}}.
\newblock Astute rag: Overcoming imperfect retrieval augmentation and knowledge conflicts for large language models.
\newblock \emph{arXiv preprint arXiv:2410.07176}.

\bibitem[{Wang et~al.(2023)Wang, Yang, and Wei}]{query2doc}
Liang Wang, Nan Yang, and Furu Wei. 2023.
\newblock Query2doc: Query expansion with large language models.
\newblock In \emph{Proceedings of the 2023 Conference on Empirical Methods in Natural Language Processing, {EMNLP} 2023, Singapore, December 6-10, 2023}, pages 9414--9423. Association for Computational Linguistics.

\bibitem[{Wei et~al.(2022)Wei, Wang, Schuurmans, Bosma, Ichter, Xia, Chi, Le, and Zhou}]{cot}
Jason Wei, Xuezhi Wang, Dale Schuurmans, Maarten Bosma, Brian Ichter, Fei Xia, Ed~H. Chi, Quoc~V. Le, and Denny Zhou. 2022.
\newblock Chain-of-thought prompting elicits reasoning in large language models.
\newblock In \emph{Advances in Neural Information Processing Systems 35: Annual Conference on Neural Information Processing Systems 2022, NeurIPS 2022, New Orleans, LA, USA, November 28 - December 9, 2022}.

\bibitem[{Xiao et~al.(2023)Xiao, Liu, Zhang, and Muennighoff}]{bge_embedding}
Shitao Xiao, Zheng Liu, Peitian Zhang, and Niklas Muennighoff. 2023.
\newblock \href {http://arxiv.org/abs/2309.07597} {C-pack: Packaged resources to advance general chinese embedding}.

\bibitem[{Xu et~al.(2024)Xu, Pang, Shen, Cheng, and Chua}]{searchain}
Shicheng Xu, Liang Pang, Huawei Shen, Xueqi Cheng, and Tat{-}Seng Chua. 2024.
\newblock Search-in-the-chain: Interactively enhancing large language models with search for knowledge-intensive tasks.
\newblock In \emph{Proceedings of the {ACM} on Web Conference 2024, {WWW} 2024, Singapore, May 13-17, 2024}, pages 1362--1373. {ACM}.

\bibitem[{Yang et~al.(2018)Yang, Qi, Zhang, Bengio, Cohen, Salakhutdinov, and Manning}]{hotpotqa}
Zhilin Yang, Peng Qi, Saizheng Zhang, Yoshua Bengio, William~W. Cohen, Ruslan Salakhutdinov, and Christopher~D. Manning. 2018.
\newblock Hotpotqa: {A} dataset for diverse, explainable multi-hop question answering.
\newblock In \emph{Proceedings of the 2018 Conference on Empirical Methods in Natural Language Processing, Brussels, Belgium, October 31 - November 4, 2018}, pages 2369--2380. Association for Computational Linguistics.

\bibitem[{Yao et~al.(2023)Yao, Yu, Zhao, Shafran, Griffiths, Cao, and Narasimhan}]{Yao-arxiv-2023-Tree}
Shunyu Yao, Dian Yu, Jeffrey Zhao, Izhak Shafran, Thomas~L. Griffiths, Yuan Cao, and Karthik Narasimhan. 2023.
\newblock Tree of thoughts: Deliberate problem solving with large language models.
\newblock \emph{arXiv preprint}.

\bibitem[{Yao et~al.(2022)Yao, Zhao, Yu, Du, Shafran, Narasimhan, and Cao}]{Yao-arxiv-2022-ReAct}
Shunyu Yao, Jeffrey Zhao, Dian Yu, Nan Du, Izhak Shafran, Karthik Narasimhan, and Yuan Cao. 2022.
\newblock React: Synergizing reasoning and acting in language models.
\newblock \emph{arXiv preprint}.

\bibitem[{Zelikman et~al.(2024)Zelikman, Harik, Shao, Jayasiri, Haber, and Goodman}]{quiet-star}
Eric Zelikman, Georges Harik, Yijia Shao, Varuna Jayasiri, Nick Haber, and Noah~D. Goodman. 2024.
\newblock \href {https://doi.org/10.48550/ARXIV.2403.09629} {Quiet-star: Language models can teach themselves to think before speaking}.
\newblock \emph{CoRR}, abs/2403.09629.

\bibitem[{Zhang et~al.(2024)Zhang, Wu, Lei, Che, Li, Xie, Huang, Zhang, Pavone, Li et~al.}]{zhang2024llama}
Di~Zhang, Jianbo Wu, Jingdi Lei, Tong Che, Jiatong Li, Tong Xie, Xiaoshui Huang, Shufei Zhang, Marco Pavone, Yuqiang Li, et~al. 2024.
\newblock Llama-berry: Pairwise optimization for o1-like olympiad-level mathematical reasoning.
\newblock \emph{arXiv preprint arXiv:2410.02884}.

\bibitem[{Zhao et~al.(2023{\natexlab{a}})Zhao, Li, Joty, Qin, and Bing}]{ZhaoLJQB23}
Ruochen Zhao, Xingxuan Li, Shafiq Joty, Chengwei Qin, and Lidong Bing. 2023{\natexlab{a}}.
\newblock Verify-and-edit: {A} knowledge-enhanced chain-of-thought framework.
\newblock In \emph{Proceedings of the 61st Annual Meeting of the Association for Computational Linguistics (Volume 1: Long Papers), {ACL} 2023, Toronto, Canada, July 9-14, 2023}, pages 5823--5840. Association for Computational Linguistics.

\bibitem[{Zhao et~al.(2023{\natexlab{b}})Zhao, Zhou, Li, Tang, Wang, Hou, Min, Zhang, Zhang, Dong, Du, Yang, Chen, Chen, Jiang, Ren, Li, Tang, Liu, Liu, Nie, and Wen}]{llm_survey}
Wayne~Xin Zhao, Kun Zhou, Junyi Li, Tianyi Tang, Xiaolei Wang, Yupeng Hou, Yingqian Min, Beichen Zhang, Junjie Zhang, Zican Dong, Yifan Du, Chen Yang, Yushuo Chen, Zhipeng Chen, Jinhao Jiang, Ruiyang Ren, Yifan Li, Xinyu Tang, Zikang Liu, Peiyu Liu, Jian{-}Yun Nie, and Ji{-}Rong Wen. 2023{\natexlab{b}}.
\newblock A survey of large language models.
\newblock \emph{CoRR}, abs/2303.18223.

\bibitem[{Zhu et~al.(2023)Zhu, Wang, Zhang, Zhang, Huang, Gan, Zhang, and Yang}]{ZhuWZZ0GZY23}
Xinyu Zhu, Junjie Wang, Lin Zhang, Yuxiang Zhang, Yongfeng Huang, Ruyi Gan, Jiaxing Zhang, and Yujiu Yang. 2023.
\newblock Solving math word problems via cooperative reasoning induced language models.
\newblock In \emph{Proceedings of the 61st Annual Meeting of the Association for Computational Linguistics (Volume 1: Long Papers), {ACL} 2023, Toronto, Canada, July 9-14, 2023}, pages 4471--4485. Association for Computational Linguistics.

\end{thebibliography}
